\documentclass[10pt,twocolumn,letterpaper]{article}

\usepackage[numbers]{natbib}

\usepackage{iccv}
\usepackage{times}
\usepackage{epsfig}
\usepackage{graphicx}
\usepackage{amsmath}
\usepackage{amssymb}
\usepackage{booktabs}
\usepackage{subcaption}

\usepackage[breaklinks=true,bookmarks=false]{hyperref}

\iccvfinalcopy 

\def\iccvPaperID{32} 

\ificcvfinal\fi

\begin{document}

\title{Towards Vision-Language Mechanistic Interpretability:\\A Causal Tracing Tool for BLIP}

\author{Vedant Palit\thanks{Corresponding authors: {\tt vedantpalit@kgpian.iitkgp.ac.in}, {\tt rohan@reworkd.ai}}\\
IIT Kharagpur
\and
Rohan Pandey$^*$\\
Reworkd.ai
\and
Aryaman Arora\\
Georgetown University
\and
Paul Pu Liang\\
Carnegie Mellon University
}





\maketitle
\ificcvfinal\thispagestyle{empty}\fi

\begin{abstract}
   Mechanistic interpretability seeks to understand the neural mechanisms that enable specific behaviors in Large Language Models (LLMs) by leveraging causality-based methods. While these approaches have identified neural circuits that copy spans of text, capture factual knowledge, and more, they remain unusable for multimodal models since adapting these tools to the vision-language domain requires considerable architectural changes. In this work, we adapt a unimodal causal tracing tool to BLIP to enable the study of the neural mechanisms underlying image-conditioned text generation. We demonstrate our approach on a visual question answering dataset, highlighting the causal relevance of later layer representations for all tokens. Furthermore, we release our BLIP causal tracing tool as open source to enable further experimentation in vision-language mechanistic interpretability by the community. Our code is available at \href{https://github.com/vedantpalit/Towards-Vision-Language-Mechanistic-Interpretability}{this URL}.
\end{abstract}

\section{Introduction}
Mechanistic interpretability \cite{olah} analyzes neural networks with the goal of reverse engineering the algorithms a network implicitly learns in their parameters. This allows for finer-grained control over a model's knowledge \cite{meng2022,meng2022mass,hernandez2023measuring} and behavior \cite{li2023circuit}. In particular, causal mediation analysis (CMA) \cite{pearl2022direct} is a popular mechanistic interpretability method that studies the effect of introducing a mediator on the outcome of a system.
However, CMA has so far been implemented only for the unimodal language domain \cite{meng2022}, limiting our understanding to this narrow class of models \cite{brown2020language}.
    

\begin{figure}
\begin{center}
   \includegraphics[width=1\linewidth]{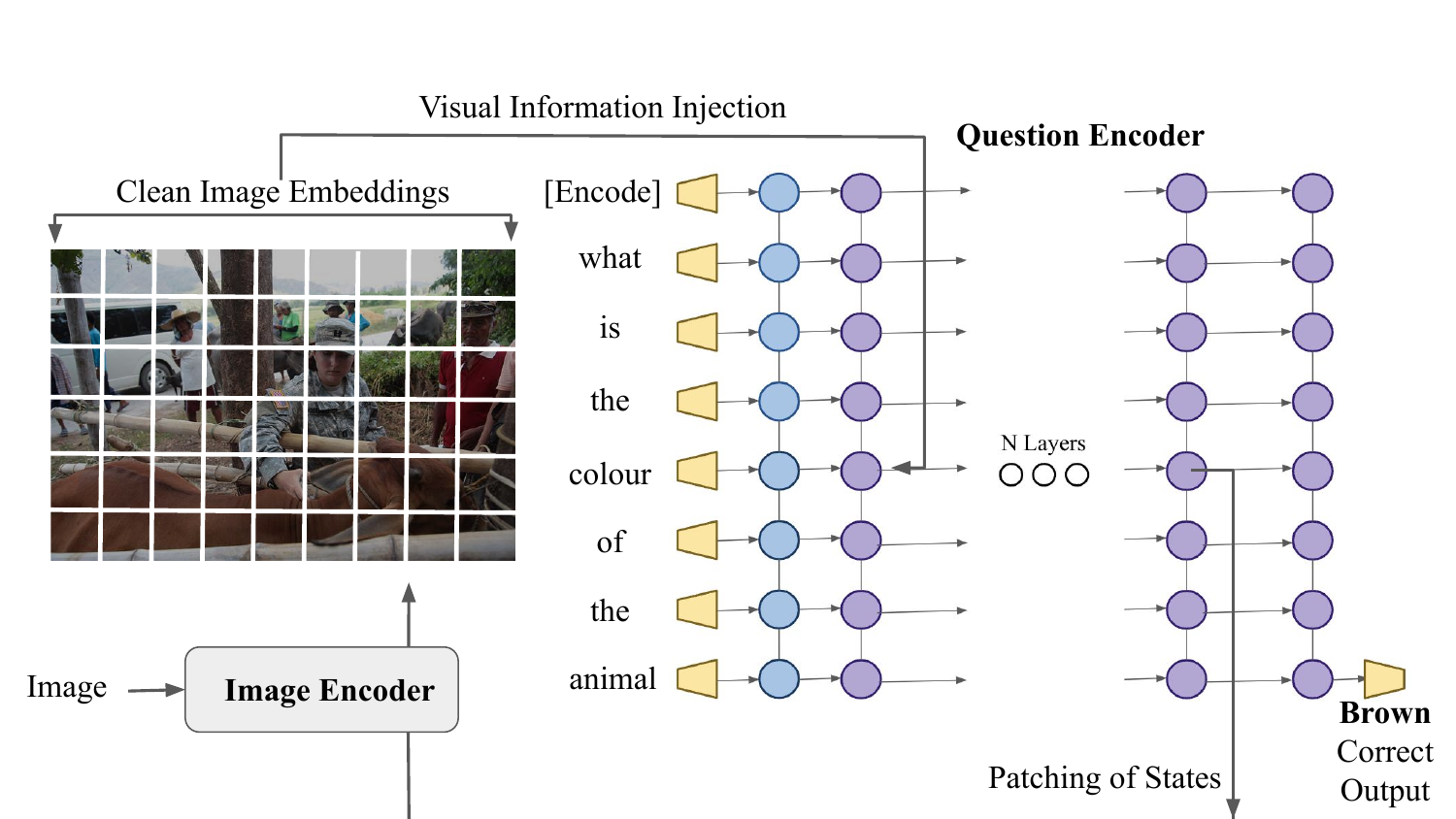}
   \includegraphics[width=1\linewidth]{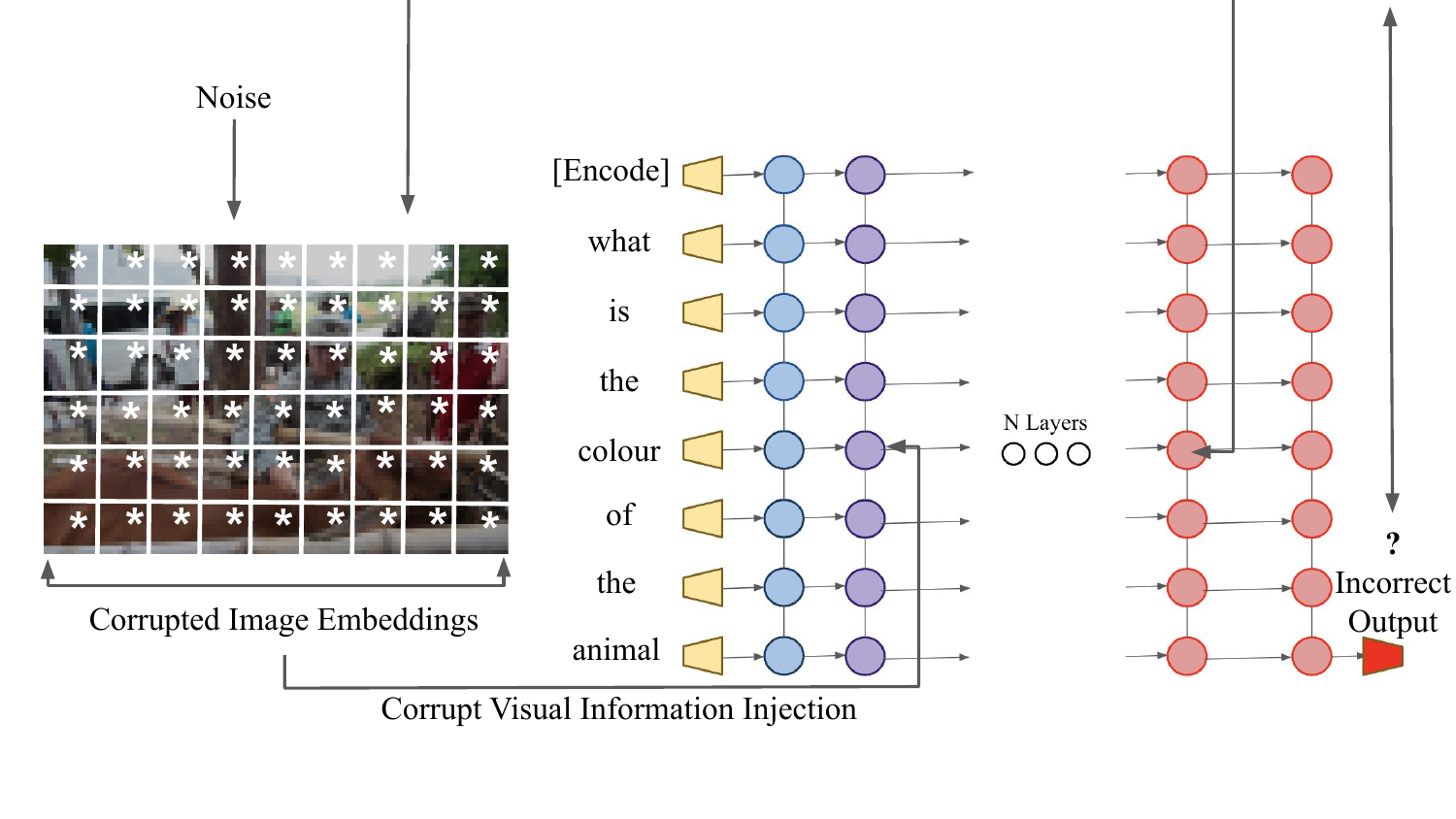}
\end{center}
\vspace{-4mm}
   \caption{
   Causal intervention to measure state's relevance:
   Above, an image of a cow is encoded, cross-attends with the question encoding, and results in the correct answer ``brown''. 
   Below, the same image encoding is corrupted, cross-attends with the question encoding, and results in an incorrect answer. 
   An intermediate state is patched from the clean to the corrupted run to observe the state's effect on the answer probabilities.}
   \label{fig:causalintro}
\end{figure}

\begin{figure}[t]
\begin{center}
   \includegraphics[width=0.85\columnwidth]{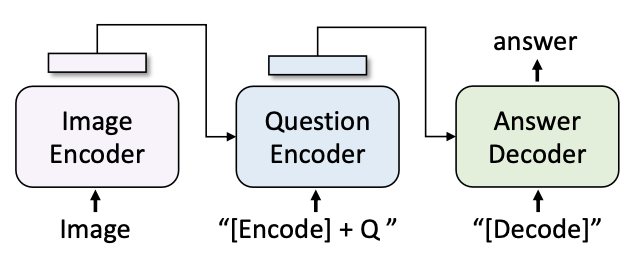}
\end{center}
   \caption{The BLIP-for-VQA \cite{li2022blip} architecture: embedding for an image patch is fed into the question encoder alongside question tokens to generate image-conditioned question embeddings through cross-attention, which are finally input to the answer decoder for answer generation.}
   \label{fig:blip}
\end{figure}

In recent years, multimodal models have rapidly grown in relevance as vision-language transformers have enabled strong performance on image-text retrieval, image captioning, and visual question answering (VQA) tasks~\cite{liang2022foundations}. Considering the powerful effects of visual stimulus on semantic representations in humans \cite{lakoff2008metaphors}, it is important to understand how similar processes occur in vision-language models. Take as an example the vision-language transformer BLIP \cite{li2022blip}, which consists of an image encoder cross-attending with a text encoder, jointly conditioning a text decoder (Fig. \ref{fig:blip}). In this work, we seek a deeper understanding of how BLIP performs VQA by adapting CMA to the vision-language setting.

\section{Related Work}
\citet{pearl2022direct} introduces causal mediation analysis by measuring the change in a response variable following an intervention, taking into consideration the effects of intermediaries or mediators.
\citet{vig2020investigating} applies this analysis to language models of the GPT-2 family to study how grammatical gender bias is mediated by the different components inside a model. They argue that probing representations \cite{adi2017finegrained,giulianelli2021hood,conneau-etal-2018-cram} for information does not tell us \cite{belinkov-glass-2019-analysis,tenney2019bert} whether the model actually uses this information, and causal approaches to interpretability are a better approach.

Meanwhile, researchers in mechanistic interpretability have developed a variety of techniques to better understand neurons and mechanisms inside neural networks (particularly unimodal language models), building on earlier work on identifying circuits in vision models \cite{olah2020zoom}. This includes applying linear algebra to understand interactions between modules inside the transformer architecture \cite{elhage2021mathematical,olsson2022context}, studying the training dynamics of transformer models, often on simple tasks \cite{elhage2022toy,nanda2022progress,bietti2023birth,chughtai2023toy,juneja2022linear}, intervening on model-internal activations to identify causal relationships between model components \cite{geiger2021,wang2022interpretability,geiger2023causal,wu2023interpretability,conmy2023towards,goldowsky2023localizing}, and attempting to map neuron features to human-interpretable concepts \cite{zimmermann2023scale,gurnee2023finding,bills2023language}.

\citet{meng2022} also base their causal intervention methods on the previous works by corrupting token embedding inputs to a language model (GPT-2 XL, GPT-NeoX) to measure causal relevance of states for capturing factual knowledge. The corruption in the input is produced by introducing noise into a sentence's subject tokens. Following this, the models are observed in three different runs---a clean input run, a corrupted input run, and an intervention involving patching of the layer outputs from a clean run of the same sentence input to the corresponding layer outputs of a corrupted run. Our implementation follows this work most closely.

On the multimodal side of interpretability, there have been thorough analyses of vision-language transformers leveraging probing approaches \cite{cao2020behind, salin2022vision}, though these face the same epistemic issues as those in the unimodal setting \cite{belinkov-glass-2019-analysis}. Another line of work explores unimodal interactions present in a multimodal model and proposes methods to understand the nature \& degree of these interactions \cite{tsai2020multimodal, liang2022multiviz}. \citet{gargi21} present a comprehensive survey of interpretability in multimodal machine learning until early 2021. Finally, \citet{kervadec2021transferable} present some interpretability experiments on transformers trained for VQA, specifically concerned with their reasoning ability.

\section{Method}

We adapt the causal intervention method from \citet{meng2022} to investigate visual question answering (COCO-QA) in a vision-language model (BLIP).

\subsection{Causal Tracing for BLIP}

\begin{table}
\small
\begin{center}
\begin{tabular}{lr}
\toprule
\textbf{Task} & \textbf{Accuracy} \\
\midrule
Color Identification & \textbf{80.23\%} \\
Location Identification & 26.30\% \\
Object Counting & 3.27\%\\
\bottomrule
\end{tabular}
\end{center}
\caption{BLIP Performance on COCO-QA Task Categories}
\end{table}

\begin{figure}
    \centering
    \begin{subfigure}[b]{0.48\columnwidth}
        \centering
        \includegraphics[width=\textwidth]{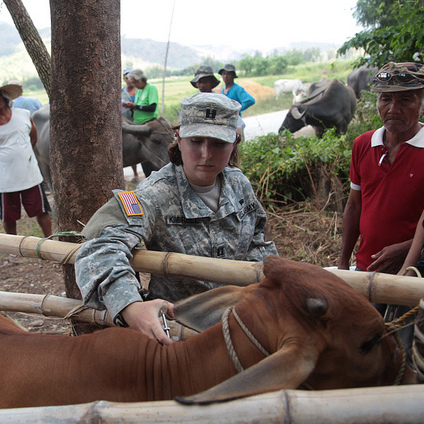}
        \caption{COCOQA-ID458864: What is the color of the animal?}
    \end{subfigure}
    \hfill
    \begin{subfigure}[b]{0.47\columnwidth}
        \centering
        \includegraphics[width=\textwidth]{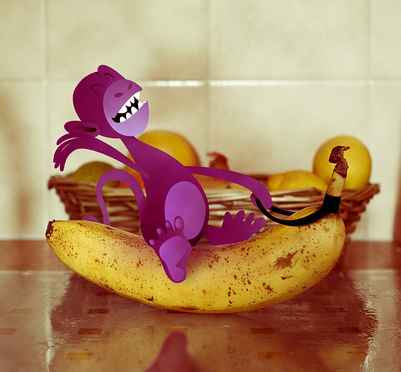}
        \caption{COCOQA-ID220218: What is the color of the character?}
    \end{subfigure}
    \caption{Two example images from COCO-QA and their accompanying questions.}
    \label{fig:cocoexamples}
\end{figure}

As input, BLIP takes a pre-processed image and question tokens, returning a single-word answer as output. We corrupt the image embeddings before they are fed into the question encoder, resulting in an incorrect output. Following this, we try and `make the answer correct again' by patching individual intermediate states (token embeddings at a layer) of a clean run into the corrupted run. The states that result in the greatest answer improvement are considered causally relevant.

\paragraph{Corruption and Patching} The second image embedding of the batch is corrupted by adding noise to all the 577 patch embeddings of the image, resulting in pairs of clean and corrupted embeddings $(E, E^*)$. For each image, we sample a single instance of noise $\epsilon \sim \mathcal{N}(1, \nu)$, where $\nu$ is an adjustable hyperparameter (standard deviation of noise), and multiply $\epsilon$ to the embedding for each patch. This corrupted image encoding is then passed into the question encoder alongside question input tokens for patching. 

To perform the causal intervention, the output of each individual state (layer $L$, token $T$) of the $E^*$ question encoder run is overwritten with the corresponding state from the clean image embedding run $E$ (see Fig. \ref{fig:causalintro}). Finally, we measure the resulting effect on output logits. This intervention process is also replicated for the answer decoder block.


\paragraph{Metrics} Given the question text embedding $\text{Q}$ and the image embedding pair $(E, E^*)$, to measure the effect of our causal intervention, we compare the correct answer's (A) probability between the corrupted run $p(\text{A} \mid E^*,\text{Q})$ and the restored run (where we patch from the clean run into the corrupted run at layer $L$ and token position $T$), and normalize across the difference between the clean and corrupted run probabilities:

\begin{equation}
    \Gamma_{L,T} = \frac{p(\text{A} \mid \text{patch}_{L, T}(E, E^*), \text{Q})-p(\text{A} \mid E^*,\text{Q})}{p(\text{A} \mid E, \text{Q})-p(\text{A} \mid E^*,\text{Q})}
\end{equation}


We expect $\Gamma_{L,T}$ to be in the range $[0, 1]$, where $0$ represents no improvement from complete corruption and $1$ represents perfect recovery of the original answer probability.

\begin{figure*}[h]
\centering
    \begin{subfigure}[b]{0.3\textwidth}
        \centering
        \includegraphics[width=1\linewidth]{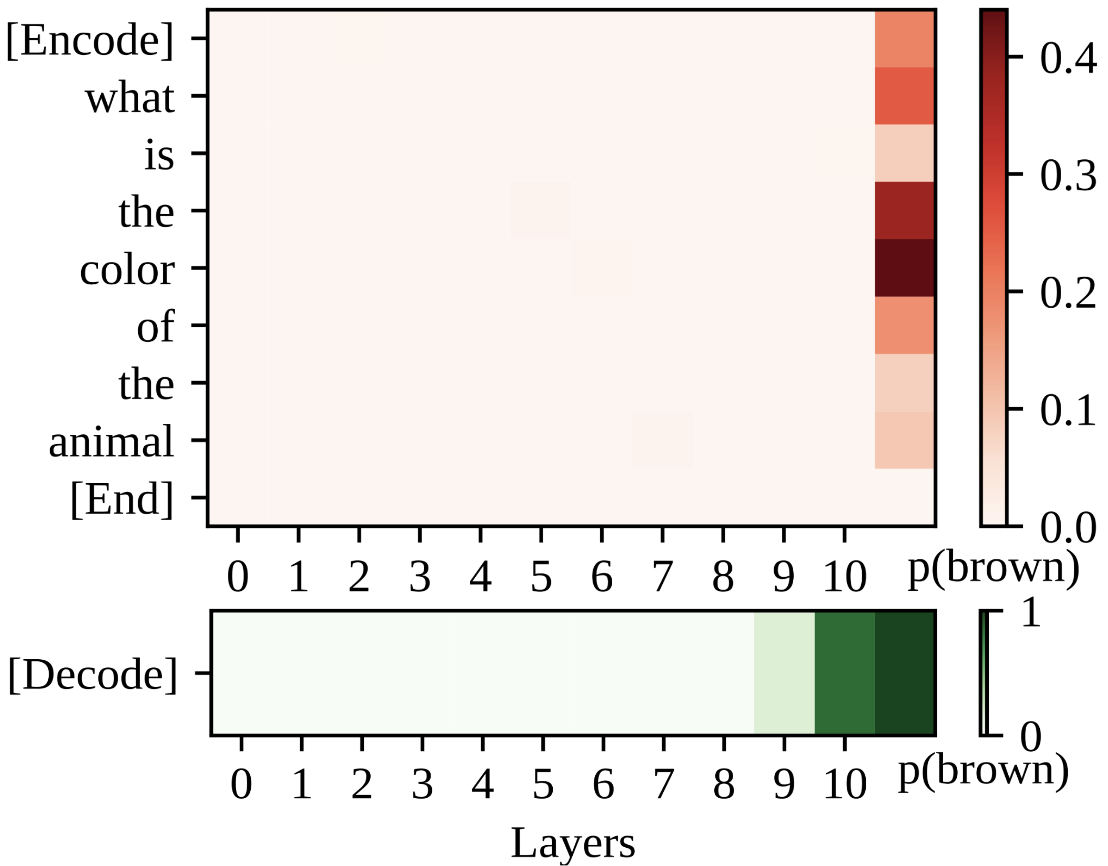}
        \caption{COCOQA-ID458864}
        \label{fig:cocoex1}
    \end{subfigure}
    \hfill
    \begin{subfigure}[b]{0.3\textwidth}
        \centering
        \includegraphics[width=1\linewidth]{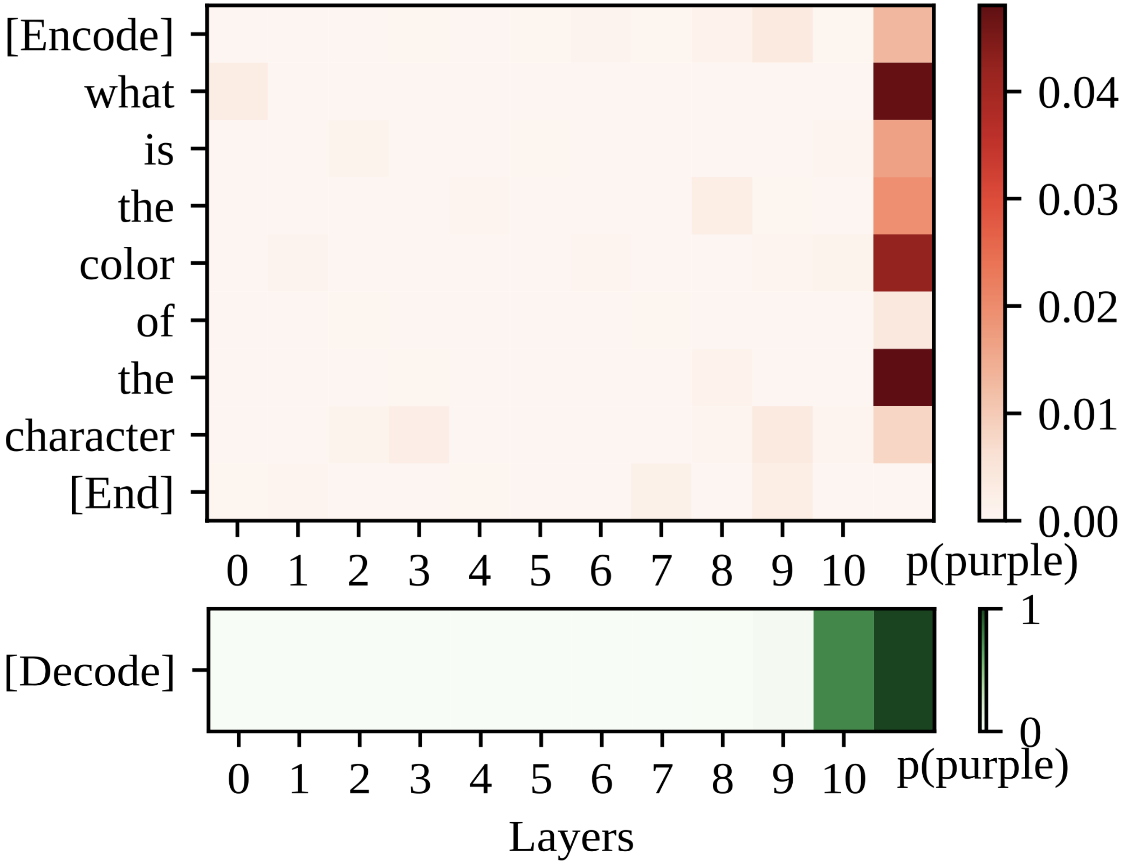}
        \caption{COCOQA-ID220218}
        \label{fig:cocoex2}
    \end{subfigure}
    \hfill
    \begin{subfigure}[b]{0.3\textwidth}
        \centering
        \includegraphics[width=1\linewidth]{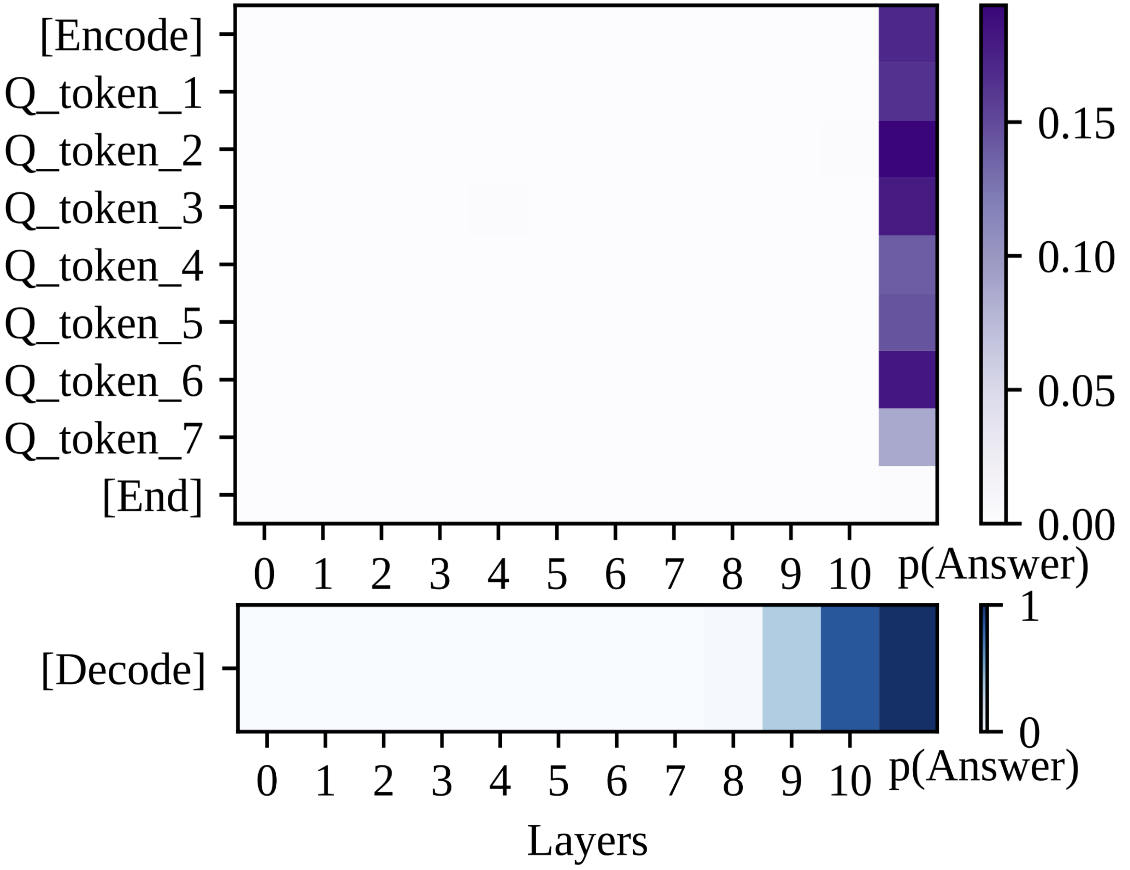}
        \caption{Average over 200 samples}
        \label{fig:coco200}
    \end{subfigure}
    \caption{Probability $\Gamma_{L,T}$ of the correct answer after performing causal interventions at specific layers on specific tokens in the question encoder (above) and answer decoder (below). Most of the causal relevance is concentrated in the final layers of the encoder as well as decoder blocks.}
    \label{fig:heatmaps}
\end{figure*}

We may then plot $\Gamma_{L,T}$ for all $(L,T)$ pairs to observe the causal relevance of that state on producing the correct answer. The darker shades of the heatmaps in Fig. \ref{fig:heatmaps} represent high causal relevance $\Gamma_{L,T}$. We can also compute an average probability difference as a function of the noise factor:

\begin{equation}
    \Gamma(\nu) = \frac{1}{|L| \cdot |T|} \sum_{l \in L} \sum_{t \in T} \Gamma_{l,t}(E^* = \nu E)
\end{equation}

We plot this function in Fig. \ref{fig:noise}, illustrating how the average difference in answer probabilities varies depending on the strength of the image embedding's corruption noise.


\subsection{COCO-QA Dataset}
VQA is an open-ended answer generation task which requires the model to predict an answer given an image and associated question input. We utilize this task as a simple testbed for causal tracing vision-language models. 
The dataset used we use is COCO-QA \cite{ren2015exploring} consisting of 123,287 images, 78,736 train and 38,948 test questions. This was sourced from MSCOCO \cite{lin2014microsoft}. The COCO-QA dataset contains one-word answers to questions belonging to four categories: object identification, object counting, colour identification, and location identification. 

We divided the training subset of COCO-QA into three splits pertaining to each of the three categories: colour, location identification, and counting. Following this division, BLIP's zero-shot performance was assessed on each of the datasets individually, results of which are shown in Table 1.

The accuracy percentages demonstrate that BLIP's pre-trained VQA model performs best in the color identification task. Further analysis showed that BLIP tends to output number of objects in an image using digits rather than natural language, which causes a low accuracy score on textual answers. Similarly, it also differs in answer structuring in the location identification task. Thus, we utilize the color identification data split of COCO-QA for causal tracing, since we want to understand mechanisms behind a behavior that a model is highly performant at.

\section{Results}

\begin{figure}
\centering
    \begin{subfigure}[b]{\columnwidth}
        \centering
        \includegraphics[width=0.88\textwidth]{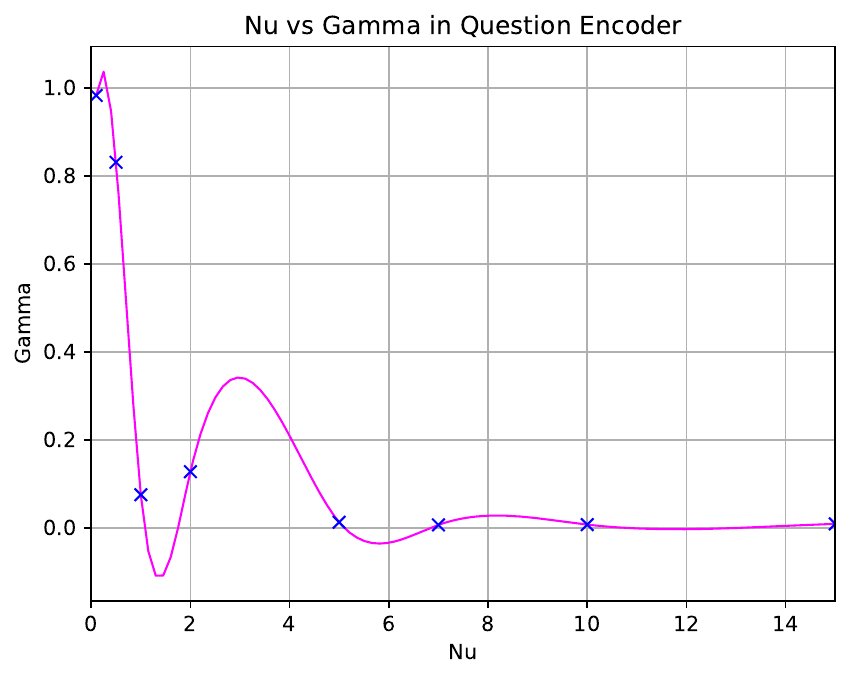}
        \caption{Question Encoder}
    \end{subfigure}
    \begin{subfigure}[b]{\columnwidth}
        \centering
        \includegraphics[width=0.88\textwidth]{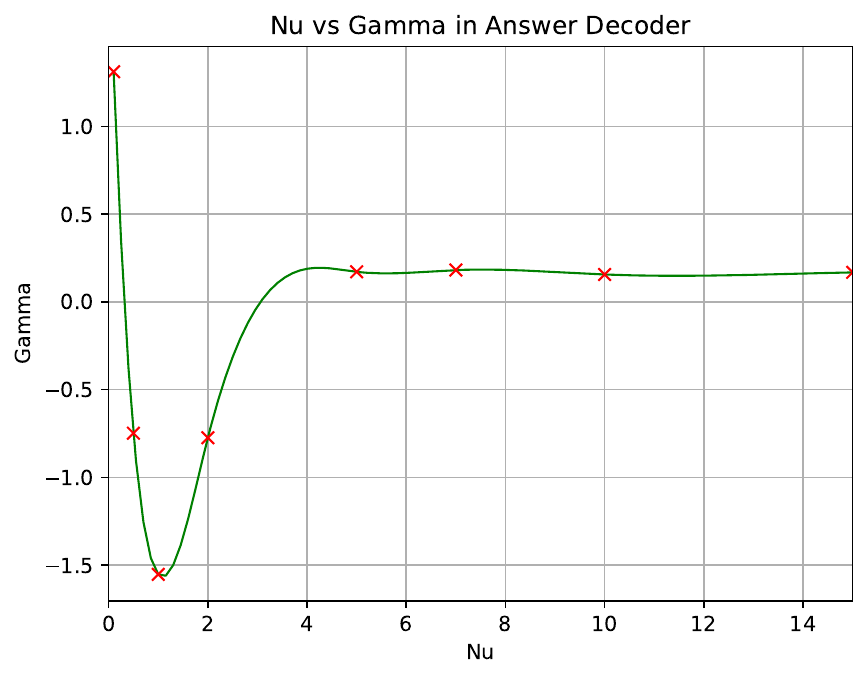}
        \caption{Answer Decoder}
    \end{subfigure}
    \caption{Effect of adjusting the noise factor $\nu$ on the answer probability difference $\Gamma$ (averaged across all $L,T$ patches) for different components of the BLIP model.}
    \label{fig:noise}
\end{figure}

In order to understand the correlation between the amount of noise injected into the image embeddings with $\Gamma(\nu)$, we first plotted the effects of adjusting the noise factor $\nu$ in the range $[0.1, 30]$, averaged over 200 samples from the dataset with 10 runs for each of the samples (see Fig. \ref{fig:noise}). We do not measure $\Gamma$ when $\nu$ is 0, since we would be patching from clean runs into clean runs, so $\Gamma(\nu) = 1$. A decaying curve is observed as the $\nu$ value increases from 0.1 to 30, with very little variation in $\Gamma(\nu)$ at extremely large values and negative values for a few values of $\nu$. Keeping both the curves in mind, we refrain from injecting too little noise that patching becomes trivial or too much noise where restoration becomes impossible, hence choosing $\nu$ as 5.


The heatmaps in Fig. \ref{fig:cocoex1} and Fig. \ref{fig:cocoex2} demonstrate the causal effects in the question encoder and answer decoder for two examples from the dataset shown in Fig. \ref{fig:cocoexamples}, averaged across 10 runs. Fig. \ref{fig:coco200} demonstrates the average effects across 200 samples from the COCO-QA dataset. The encoder and decoder layers are indexed from 0--11, and input question tokens are plotted vertically.

It is clear from the figures that in the question encoder, only the final layer (11) for all tokens plays a significant role in affecting the output to a higher degree than any preceding layers or tokens. In the case of the answer decoder, the final layers (9 to 11) play the most apparent role in the final output of the model.
These results show that BLIP does not benefit from restored access to the correct image embeddings until the final few layers. This may mean that the vision modality is not relevant to model computations until the final layer, i.e. vision and language are processed independently in the intermediate layers. On the other hand, it may also mean that the final layers override preceding layers, which may still be weakly causally relevant to the model output.




\section{Conclusion}

We introduced the first causal tracing tool for a vision-language model and studied how model performance is localized in BLIP on a subset of the visual question answering task with the COCO-QA dataset. Previous work on interpretability of vision-language models has not focused on identifying causal mechanisms, so we hope that this work invigorates research in this area. Towards this end, we fully open source our code and will soon release a visualizer as well as adaptations to other vision-language models.

Many aspects of the causal tracing methodology are still not fully understood. For example, since the role of the noise factor $\nu$ is unclear, future work could study why different components of the model have different sensitivities to noise; for example, why is performance not monotonically reduced by increasing $\nu$? Also, restoration of the clean image embedding at \textit{multiple} points (instead of just one) may help us understand cross-module coordination within the model.

A bigger project is to identify larger mechanisms within vision-language models that can explain how the model performs specific tasks, as has been done in unimodal language models \cite{nanda2022progress,wang2022interpretability}. This will help us understand how multimodal models work and let us verify whether they perform tasks as expected, e.g.~whether they learn good algorithms or poor shortcuts on compositional understanding benchmarks like Winoground \cite{thrush2022winoground,pandey2023semantic}. Overall, much work remains in this line of research and we look forward to using causal intervention methods for disentangling the mechanisms learned by vision-language models.





{\small
\bibliographystyle{ieee_fullname}
\bibliography{egpaper_final}

\begin{thebibliography}{10}\itemsep=-1pt

\bibitem[Adi et~al.(2017)]{adi2017finegrained}
Yossi Adi, Einat Kermany, Yonatan Belinkov, Ofer Lavi, and Yoav Goldberg.
\newblock Fine-grained analysis of sentence embeddings using auxiliary
  prediction tasks, 2017.

\bibitem[Belinkov and Glass(2019)]{belinkov-glass-2019-analysis}
Yonatan Belinkov and James Glass.
\newblock Analysis methods in neural language processing: A survey.
\newblock {\em Transactions of the Association for Computational Linguistics},
  7:49--72, 2019.

\bibitem[Bietti et~al.(2023)]{bietti2023birth}
Alberto Bietti, Vivien Cabannes, Diane Bouchacourt, Herve Jegou, and Leon
  Bottou.
\newblock Birth of a transformer: A memory viewpoint.
\newblock {\em arXiv preprint arXiv:2306.00802}, 2023.

\bibitem[Bills et~al.(2023)]{bills2023language}
Steven Bills, Nick Cammarata, Dan Mossing, Henk Tillman, Leo Gao, Gabriel Goh,
  Ilya Sutskever, Jan Leike, Jeff Wu, and William Saunders.
\newblock Language models can explain neurons in language models.
\newblock {\em OpenAI}, 2023.

\bibitem[Brown et~al.(2020)]{brown2020language}
Tom~B. Brown, Benjamin Mann, Nick Ryder, Melanie Subbiah, Jared Kaplan,
  Prafulla Dhariwal, Arvind Neelakantan, Pranav Shyam, Girish Sastry, Amanda
  Askell, Sandhini Agarwal, Ariel Herbert-Voss, Gretchen Krueger, Tom Henighan,
  Rewon Child, Aditya Ramesh, Daniel~M. Ziegler, Jeffrey Wu, Clemens Winter,
  Christopher Hesse, Mark Chen, Eric Sigler, Mateusz Litwin, Scott Gray,
  Benjamin Chess, Jack Clark, Christopher Berner, Sam McCandlish, Alec Radford,
  Ilya Sutskever, and Dario Amodei.
\newblock Language models are few-shot learners, 2020.

\bibitem[Cao et~al.(2020)]{cao2020behind}
Jize Cao, Zhe Gan, Yu Cheng, Licheng Yu, Yen-Chun Chen, and Jingjing Liu.
\newblock Behind the scene: Revealing the secrets of pre-trained
  vision-and-language models.
\newblock In {\em Computer Vision--ECCV 2020: 16th European Conference,
  Glasgow, UK, August 23--28, 2020, Proceedings, Part VI 16}, pages 565--580.
  Springer, 2020.

\bibitem[Chughtai et~al.(2023)]{chughtai2023toy}
Bilal Chughtai, Lawrence Chan, and Neel Nanda.
\newblock A toy model of universality: Reverse engineering how networks learn
  group operations.
\newblock {\em arXiv preprint arXiv:2302.03025}, 2023.

\bibitem[Conmy et~al.(2023)]{conmy2023towards}
Arthur Conmy, Augustine~N Mavor-Parker, Aengus Lynch, Stefan Heimersheim, and
  Adri{\`a} Garriga-Alonso.
\newblock Towards automated circuit discovery for mechanistic interpretability.
\newblock {\em arXiv preprint arXiv:2304.14997}, 2023.

\bibitem[Conneau et~al.(2018)]{conneau-etal-2018-cram}
Alexis Conneau, German Kruszewski, Guillaume Lample, Lo{\"\i}c Barrault, and
  Marco Baroni.
\newblock What you can cram into a single {\$}{\&}!{\#}* vector: Probing
  sentence embeddings for linguistic properties.
\newblock In {\em Proceedings of the 56th Annual Meeting of the Association for
  Computational Linguistics (Volume 1: Long Papers)}, pages 2126--2136,
  Melbourne, Australia, July 2018. Association for Computational Linguistics.

\bibitem[Elhage et~al.(2022)]{elhage2022toy}
Nelson Elhage, Tristan Hume, Catherine Olsson, Nicholas Schiefer, Tom Henighan,
  Shauna Kravec, Zac Hatfield-Dodds, Robert Lasenby, Dawn Drain, Carol Chen,
  et~al.
\newblock Toy models of superposition.
\newblock {\em arXiv preprint arXiv:2209.10652}, 2022.

\bibitem[Elhage et~al.(2021)]{elhage2021mathematical}
Nelson Elhage, Neel Nanda, Catherine Olsson, Tom Henighan, Nicholas Joseph, Ben
  Mann, Amanda Askell, Yuntao Bai, Anna Chen, Tom Conerly, et~al.
\newblock A mathematical framework for transformer circuits.
\newblock {\em Transformer Circuits Thread}, 2021.

\bibitem[Geiger et~al.(2021)]{geiger2021}
Atticus Geiger, Hanson Lu, Thomas Icard, and Christopher Potts.
\newblock Causal abstractions of neural networks.
\newblock In M. Ranzato, A. Beygelzimer, Y. Dauphin, P.S. Liang, and J.~Wortman
  Vaughan, editors, {\em Advances in Neural Information Processing Systems},
  volume~34, pages 9574--9586. Curran Associates, Inc., 2021.

\bibitem[Geiger et~al.(2023)]{geiger2023causal}
Atticus Geiger, Chris Potts, and Thomas Icard.
\newblock Causal abstraction for faithful model interpretation.
\newblock {\em arXiv preprint arXiv:2301.04709}, 2023.

\bibitem[Giulianelli et~al.(2021)]{giulianelli2021hood}
Mario Giulianelli, Jacqueline Harding, Florian Mohnert, Dieuwke Hupkes, and
  Willem Zuidema.
\newblock Under the hood: Using diagnostic classifiers to investigate and
  improve how language models track agreement information, 2021.

\bibitem[Goldowsky-Dill et~al.(2023)]{goldowsky2023localizing}
Nicholas Goldowsky-Dill, Chris MacLeod, Lucas Sato, and Aryaman Arora.
\newblock Localizing model behavior with path patching.
\newblock {\em arXiv preprint arXiv:2304.05969}, 2023.

\bibitem[Gurnee et~al.(2023)]{gurnee2023finding}
Wes Gurnee, Neel Nanda, Matthew Pauly, Katherine Harvey, Dmitrii Troitskii, and
  Dimitris Bertsimas.
\newblock Finding neurons in a haystack: Case studies with sparse probing.
\newblock {\em arXiv preprint arXiv:2305.01610}, 2023.

\bibitem[Hernandez et~al.(2023)]{hernandez2023measuring}
Evan Hernandez, Belinda~Z Li, and Jacob Andreas.
\newblock Measuring and manipulating knowledge representations in language
  models.
\newblock {\em arXiv preprint arXiv:2304.00740}, 2023.

\bibitem[Joshi et~al.(2021)]{gargi21}
Gargi Joshi, Rahee Walambe, and Ketan Kotecha.
\newblock A review on explainability in multimodal deep neural nets.
\newblock {\em IEEE Access}, 9:59800--59821, 2021.

\bibitem[Juneja et~al.(2022)]{juneja2022linear}
Jeevesh Juneja, Rachit Bansal, Kyunghyun Cho, Jo{\~a}o Sedoc, and Naomi Saphra.
\newblock Linear connectivity reveals generalization strategies.
\newblock {\em arXiv preprint arXiv:2205.12411}, 2022.

\bibitem[Kervadec et~al.(2021)]{kervadec2021transferable}
Corentin Kervadec, Theo Jaunet, Grigory Antipov, Moez Baccouche, Romain
  Vuillemot, and Christian Wolf.
\newblock How transferable are reasoning patterns in vqa?
\newblock In {\em Proceedings of the IEEE/CVF Conference on Computer Vision and
  Pattern Recognition}, pages 4207--4216, 2021.

\bibitem[Lakoff and Johnson(2008)]{lakoff2008metaphors}
George Lakoff and Mark Johnson.
\newblock {\em Metaphors we live by}.
\newblock University of Chicago press, 2008.

\bibitem[Li et~al.(2022)]{li2022blip}
Junnan Li, Dongxu Li, Caiming Xiong, and Steven Hoi.
\newblock {BLIP}: Bootstrapping language-image pre-training for unified
  vision-language understanding and generation.
\newblock In {\em International Conference on Machine Learning}, pages
  12888--12900. PMLR, 2022.

\bibitem[Li et~al.(2023)]{li2023circuit}
Maximilian Li, Xander Davies, and Max Nadeau.
\newblock Circuit breaking: Removing model behaviors with targeted ablation.
\newblock In {\em DeployableGenerativeAI}, 2023.

\bibitem[Liang et~al.(2022)]{liang2022multiviz}
Paul~Pu Liang, Yiwei Lyu, Gunjan Chhablani, Nihal Jain, Zihao Deng, Xingbo
  Wang, Louis-Philippe Morency, and Ruslan Salakhutdinov.
\newblock Multiviz: Towards visualizing and understanding multimodal models.
\newblock In {\em The Eleventh International Conference on Learning
  Representations}, 2022.

\bibitem[Liang et~al.(2022)]{liang2022foundations}
Paul~Pu Liang, Amir Zadeh, and Louis-Philippe Morency.
\newblock Foundations and recent trends in multimodal machine learning:
  Principles, challenges, and open questions.
\newblock {\em arXiv preprint arXiv:2209.03430}, 2022.

\bibitem[Lin et~al.(2014)]{lin2014microsoft}
Tsung-Yi Lin, Michael Maire, Serge Belongie, James Hays, Pietro Perona, Deva
  Ramanan, Piotr Doll{\'a}r, and C~Lawrence Zitnick.
\newblock Microsoft {COCO}: Common objects in context.
\newblock In {\em Computer Vision--ECCV 2014: 13th European Conference, Zurich,
  Switzerland, September 6-12, 2014, Proceedings, Part V 13}, pages 740--755.
  Springer, 2014.

\bibitem[Meng et~al.(2022)]{meng2022}
Kevin Meng, David Bau, Alex Andonian, and Yonatan Belinkov.
\newblock Locating and editing factual associations in {GPT}.
\newblock In S. Koyejo, S. Mohamed, A. Agarwal, D. Belgrave, K. Cho, and A. Oh,
  editors, {\em Advances in Neural Information Processing Systems}, volume~35,
  pages 17359--17372. Curran Associates, Inc., 2022.

\bibitem[Meng et~al.(2022)]{meng2022mass}
Kevin Meng, Arnab~Sen Sharma, Alex Andonian, Yonatan Belinkov, and David Bau.
\newblock Mass-editing memory in a transformer.
\newblock {\em arXiv preprint arXiv:2210.07229}, 2022.

\bibitem[Nanda et~al.(2023)]{nanda2022progress}
Neel Nanda, Lawrence Chan, Tom Lieberum, Jess Smith, and Jacob Steinhardt.
\newblock Progress measures for grokking via mechanistic interpretability.
\newblock In {\em The Eleventh International Conference on Learning
  Representations}, 2023.

\bibitem[Olah(2022)]{olah}
Chris Olah.
\newblock Mechanistic interpretability, variables, and the importance of
  interpretable bases.
\newblock {\em Transformer Circuits Thread}, 2022.

\bibitem[Olah et~al.(2020)]{olah2020zoom}
Chris Olah, Nick Cammarata, Ludwig Schubert, Gabriel Goh, Michael Petrov, and
  Shan Carter.
\newblock Zoom in: An introduction to circuits.
\newblock {\em Distill}, 5(3):e00024--001, 2020.

\bibitem[Olsson et~al.(2022)]{olsson2022context}
Catherine Olsson, Nelson Elhage, Neel Nanda, Nicholas Joseph, Nova DasSarma,
  Tom Henighan, Ben Mann, Amanda Askell, Yuntao Bai, Anna Chen, et~al.
\newblock In-context learning and induction heads.
\newblock {\em arXiv preprint arXiv:2209.11895}, 2022.

\bibitem[Pandey(2023)]{pandey2023semantic}
Rohan Pandey.
\newblock Semantic composition in visually grounded language models.
\newblock {\em arXiv preprint arXiv:2305.16328}, 2023.

\bibitem[Pearl(2022)]{pearl2022direct}
Judea Pearl.
\newblock Direct and indirect effects.
\newblock In {\em Probabilistic and causal inference: the works of Judea
  Pearl}, pages 373--392. 2022.

\bibitem[Ren et~al.(2015)]{ren2015exploring}
Mengye Ren, Ryan Kiros, and Richard Zemel.
\newblock Exploring models and data for image question answering.
\newblock {\em Advances in neural information processing systems}, 28, 2015.

\bibitem[Salin et~al.(2022)]{salin2022vision}
Emmanuelle Salin, Badreddine Farah, St{\'e}phane Ayache, and Benoit Favre.
\newblock Are vision-language transformers learning multimodal representations?
  a probing perspective.
\newblock In {\em Proceedings of the AAAI Conference on Artificial
  Intelligence}, volume~36, pages 11248--11257, 2022.

\bibitem[Tenney et~al.(2019)]{tenney2019bert}
Ian Tenney, Dipanjan Das, and Ellie Pavlick.
\newblock {BERT} rediscovers the classical {NLP} pipeline, 2019.

\bibitem[Thrush et~al.(2022)]{thrush2022winoground}
Tristan Thrush, Ryan Jiang, Max Bartolo, Amanpreet Singh, Adina Williams, Douwe
  Kiela, and Candace Ross.
\newblock Winoground: Probing vision and language models for visio-linguistic
  compositionality.
\newblock In {\em Proceedings of the IEEE/CVF Conference on Computer Vision and
  Pattern Recognition}, pages 5238--5248, 2022.

\bibitem[Tsai et~al.(2020)]{tsai2020multimodal}
Yao-Hung~Hubert Tsai, Martin~Q Ma, Muqiao Yang, Ruslan Salakhutdinov, and
  Louis-Philippe Morency.
\newblock Multimodal routing: Improving local and global interpretability of
  multimodal language analysis.
\newblock In {\em Proceedings of the Conference on Empirical Methods in Natural
  Language Processing. Conference on Empirical Methods in Natural Language
  Processing}, volume 2020, page 1823. NIH Public Access, 2020.

\bibitem[Vig et~al.(2020)]{vig2020investigating}
Jesse Vig, Sebastian Gehrmann, Yonatan Belinkov, Sharon Qian, Daniel Nevo,
  Yaron Singer, and Stuart Shieber.
\newblock Investigating gender bias in language models using causal mediation
  analysis.
\newblock {\em Advances in neural information processing systems},
  33:12388--12401, 2020.

\bibitem[Wang et~al.(2023)]{wang2022interpretability}
Kevin~Ro Wang, Alexandre Variengien, Arthur Conmy, Buck Shlegeris, and Jacob
  Steinhardt.
\newblock Interpretability in the wild: a circuit for indirect object
  identification in {GPT}-2 {S}mall.
\newblock In {\em The Eleventh International Conference on Learning
  Representations}, 2023.

\bibitem[Wu et~al.(2023)]{wu2023interpretability}
Zhengxuan Wu, Atticus Geiger, Christopher Potts, and Noah~D. Goodman.
\newblock Interpretability at scale: Identifying causal mechanisms in {A}lpaca.
\newblock {\em arXiv preprint arXiv:2305.08809}, 2023.

\bibitem[Zimmermann et~al.(2023)]{zimmermann2023scale}
Roland~S Zimmermann, Thomas Klein, and Wieland Brendel.
\newblock Scale alone does not improve mechanistic interpretability in vision
  models.
\newblock {\em arXiv preprint arXiv:2307.05471}, 2023.

\end{thebibliography}
}

\end{document}